\documentclass[runningheads]{llncs}

\usepackage{eccv}

\usepackage{eccvabbrv}

\usepackage{graphicx}
\usepackage{booktabs}
\usepackage{upgreek}
\usepackage{bm}
\usepackage[accsupp]{axessibility}  

\usepackage[pagebackref,breaklinks,colorlinks,citecolor=eccvblue]{hyperref}

\usepackage{orcidlink}

\usepackage{makecell}
\usepackage{colortbl}
\usepackage{xspace}
\usepackage{tabu}
\usepackage{amssymb}
\newlength\savewidth

\usepackage{enumitem}
\usepackage{amssymb}
\usepackage{pifont}
\newcommand{\cmark}{\ding{51}}%
\newcommand{\xmark}{\ding{55}}%

\newcommand\mypara[1]{\vspace{1.0mm}\noindent\textbf{#1}}

\newcommand{\ourmethod}{{\sc {GHG}}\xspace}

\newcommand{\figref}[1]{Fig.~\ref{#1}}

\newcommand{\tabref}[1]{Tab.~\ref{#1}}
\newcommand{\eqnref}[1]{Eq.~(\ref{#1})}

\begin{document}

\title{ Generalizable Human Gaussians \\ for Sparse View Synthesis}
\titlerunning{Generalizable Human Gaussians (GHG)}

\author{
Youngjoong Kwon\inst{1} 
\and
Baole Fang\inst{1}$^*$
\and
Yixing Lu\inst{1}$^*$
\and
Haoye Dong\inst{1} 
\and
Cheng Zhang\inst{1} 
\and
Francisco Vicente Carrasco\inst{1} 
\and
Albert Mosella-Montoro\inst{1} 
\and
Jianjin Xu\inst{1} 
\and \\
Shingo Takagi\inst{2} 
\and
Daeil Kim\inst{2} 
\and
Aayush Prakash\inst{2} 
\and
Fernando De la Torre\inst{1} 
}

\authorrunning{Kwon et al.}

\institute{
$^1$Carnegie Mellon University \quad
$^2$Meta Reality Labs \\
\smallskip\smallskip
\url{https://humansensinglab.github.io/Generalizable-Human-Gaussians/}
}

\maketitle

\def\thefootnote{*}\footnotetext{Equal contribution}
\begin{abstract}

Recent progress in neural rendering has brought forth pioneering methods, such as NeRF and Gaussian Splatting, which revolutionize view rendering across various domains like AR/VR, gaming, and content creation. While these methods excel at interpolating {\em within the training data}, the challenge of generalizing to new scenes and objects from very sparse views persists. Specifically, modeling 3D humans from sparse views presents formidable hurdles due to the inherent complexity of human geometry, resulting in inaccurate reconstructions of geometry and textures.
To tackle this challenge, this paper leverages recent advancements in Gaussian Splatting and introduces a new method to learn generalizable human Gaussians that allows photorealistic and accurate view-rendering of a new human subject from a limited set of sparse views in a feed-forward manner.
A pivotal innovation of our approach involves reformulating the learning of 3D Gaussian parameters into a regression process defined on the 2D UV space of a human template, which allows leveraging the strong geometry prior and the advantages of 2D convolutions. In addition, a multi-scaffold is proposed to effectively represent the offset details.
Our method outperforms recent methods on both within-dataset generalization as well as cross-dataset generalization settings.

\end{abstract}
\section{Introduction} \label{sec:intro}

\begin{figure}[t]
    \centerline{\includegraphics[width=1\linewidth]{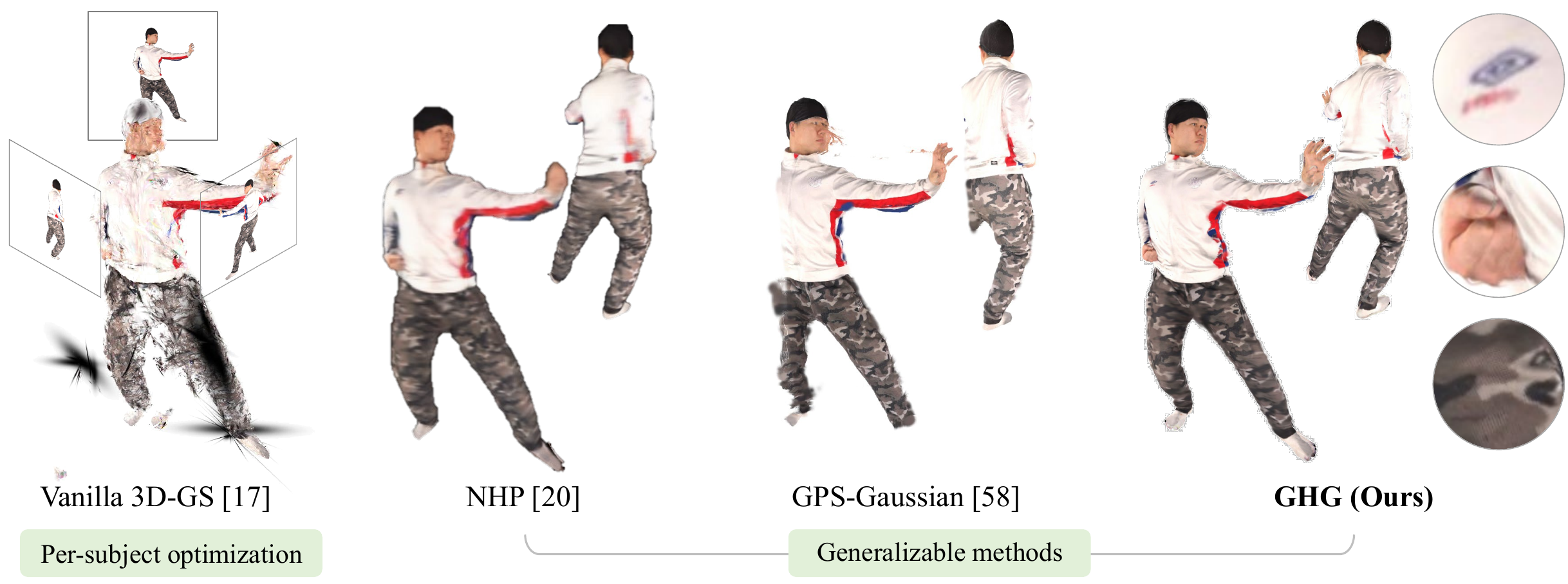}}
    \caption{\small \textbf{Generalizable Human Gaussian (GHG).} Our method can perform accurate and photorealistic novel view renderings of a new human subject given very sparse inputs (e.g., 3 views) without involving any test-time optimization or fine-tuning. In the sparse-view setup, our GHG approach exhibits superior rendering quality compared to other generalizable methods such as NHP~\cite{kwon2021neural} and GPS-Gaussian~\cite{zheng2023gps}. 
    }
    \label{fig:teaser} \vspace{-8mm}
\end{figure}

Recent advancements in neural rendering techniques, such as Neural Radiance Fields (NeRF)~\cite{mildenhall2021nerf}, 3D Gaussian Splatting~\cite{kerbl20233d}, and point-based graphics~\cite{aliev2020neural},  have unveiled a multitude of captivating applications spanning 
virtual avatars, asset content creation, or cinematic production. 
 While these methods excel at interpolating a single scene/object from many input views, it is very challenging to generalize to new scenes and objects with few samples, and extrapolating outside the captured views. This limitation is particularly pronounced in the task of photorealistic rendering of humans, a subject of widespread interest and applications. The task of modeling 3D humans from sparse viewpoints is complicated by the inherent complexities of human geometry, including articulations, self-occlusions, and complex surface geometries like hair. These factors often lead to significant inaccuracies in the reconstruction of both geometry and textures, posing substantial hurdles to generating photorealistic digital humans.

Recent advances in human rendering incorporate implicit neural representations (\eg NeRF) with human template models to facilitate generalizable and robust synthesis under sparse view settings~\cite{kwon2021neural,zhao2022humannerf,cheng2022generalizable,chen2022geometry,kwon2023neural,pan2023transhuman}.
While NeRF-based methods have made sigificant progress in generalizable human rendering, they are limited by their slow runtime, mainly due to their computationally intesive per-pixel volume rendering process. Additionally, in sparse-view setting, leveraging recent advances in inpainting models holds promiss for capturing details absent in the input views. However, integrating these modules presents practical challenges as it would further burden the already heavy NeRF-based system.

Recently, explicit representation methods such as 3D Gaussians have gained popularity for their efficient rasterization-based rendering speed. This fast rendering capability enables seamless integration with other models, such as generative~\cite{tang2023dreamgaussian} or depth estimation models~\cite{zheng2023gps}, to achieve high-quality novel view rendering. However, these methods encounter difficulties when dealing with human subjects particularly when only sparse input views (e.g., 2-3 views) are available. The inherent challenges in rendering human subjects, such as articulations, self-occlusions, and complex surface geometries, worsen the difficulties in such sparse-view setting {(see Figure~\ref{fig:teaser})}.

To this end, we propose \textbf{Generalizable Human Gaussians} (\ourmethod), a method for accurate and photorealistic novel-view renderings of human subjects. \ourmethod enables rendering of a novel human subject given very sparse input views, without requiring any test-time optimization or fine-tuning. 
To improve performance in the sparse-view setting, our key insight is to leverage human geometry prior by reformulating the optimization of 3D Gaussian parameters within the 2D UV space derived from a human template model.
By anchoring the Gaussian parameters onto the surface of the 3D human template model, each location in the template space can be mapped to each foreground pixel in the corresponding 2D UV map space. Our UV map-based Gaussian representation significantly improves the reconstruction of complex human geometries. Operating on the 2D UV map space enables us to utilize 2D CNNs for the Gaussian optimization which can incorporate information from neighboring pixels unlike MLPs. Additionally, this approach makes our model compatible with inpainting models~\cite{yu2018generative,yu2018free}, facilitating seamless integration.

While our human UV map-based representation brings significant advancement in generalization and robustness in rendering, we aim to improve its effectiveness further. Given the inherent disparity between the template body model and real human geometry (\eg clothing or hair), we present a method to bridge this gap. To achieve this, we propose to generate multiple offset meshes through dilating the human template mesh, both at the input and output spaces. These meshes serve as scaffolds, enabling effective encoding of input geometry information, as well as facilitating a richer representation of displacements beyond that can be captured by a single template mesh at the output. Leveraging these multi-scaffold meshes enables more faithful capturing of real human geometries, which often cannot be accurately represented by a single template mesh surface.

We evaluate the efficacy of our Generalizable Human Gaussians on two multi-view human capture datasets: THuman 2.0~\cite{thuman} and RenderPeople~\cite{renderpeople}. 
Existing generalizable human rendering approaches that allow sparse-view input (3 views) are primarily NeRF-based methods~\cite{kwon2021neural,chen2022geometry}.
We compare with these methods and demonstrate superior rendering quality in both in-domain and cross-dataset evaluation settings.
Additionally, we compare our approach with existing 3D Gaussians-based methods, which either necessitates more input views~\cite{zheng2023gps} or per-subject optimization~\cite{kerbl20233d}, showcasing distinct benefits of our approach.

\smallskip\smallskip
In summary, our main contributions are as follows:
\begin{itemize}[itemsep=5pt,topsep=5pt,leftmargin=15pt]
\item We propose a new feed-forward method for accurate and photorealistic novel-view renderings of new humans from very sparse input views. This is achieved by integrating human geometry prior with 3D Gaussians. Specifically, we reformulate the optimization of 3D Gaussian parameters into a task of generating a Gaussian parameter map within the 2D human UV space derived from the human template model.
\item We propose a multi-scaffold representation aimed at minimizing the disparity between the template model and real human geometry. This approach enables the Gaussian parameters to be learned across multiple scaffold spaces, allowing for a more comprehensive representation of displacements that surpasses the capacity of a single template mesh space.

\end{itemize}

\section{Related Work} \label{sec:related}

\mypara{Generalizable NeRF for Human Rendering.}
Neural Radiance Fields has demonstrated its powerful capability to render 3D scenes with photorealistic quality. 
However, they can be only optimized on a single scene, and require images taken from densely sampled cameras to train. To generalize to new scenes without optimization at inference time, some works condition the generation on the pixel-aligned features~\cite{saito2019pifu,yu2021pixelnerf,wang2022attention}, cost-volumes~\cite{chen2021mvsnerf,wang2024learning}, or image-based rendering~\cite{wang2021ibrnet,liu2022neuray}. Although they have demonstrated high-quality generalization ability on general objects and scenes, directly applying those methods to human subjects is non-trivial due to the complicated human geometries (i.e., articulations and self-occlusions).
To effectively address the generalization to humans, a line of works utilize 3D human prior. Specifically, SMPL surface~\cite{loper2023smpl} is leveraged as the tool for aggregating the relevant features while preserving its geometric structure~\cite{kwon2021neural,zhao2022humannerf,cheng2022generalizable,chen2022geometry,gao2022mps,kwon2023neural,chen2023gm,gao2023neural}. Skeletal keypoints are also utilized~\cite{mihajlovic2022keypointnerf}. Despite their detailed output, their rendering speed is very slow due to the volume rendering process which requires heavy computations to render a single pixel. This deters them from combining with other modules to further improve the performance (\eg inpainting).
In our work, thanks to the fast rasterization-based rendering, our model can be combined with 2D-based inpainting module~\cite{yu2018free,yu2018generative} to compensate for the unobserved regions inevitable under sparse view settings.

\mypara{3D Gaussian Splatting.}
3D Gaussian Splatting is a method to represent a scene with a set of 3D Gaussians~\cite{Robertini:2016,kerbl20233d}. By utilizing GPU-parallelized rasterization, they achieve fast rendering speed and have presented impressive ability in novel view synthesis tasks.  
Some concurrent works utilize human template as the 3D prior and combine it with 3D Gaussians to create animatable representations~\cite{zielonka2023drivable,zhu2023ash,moreau2023human,kocabas2023hugs,jena2023splatarmor,li2023animatable,hu2023gaussianavatar,ye2023animatable,zhou2024headstudio}. However, they are not generalizable and require new training process for every new subject. 
Zheng et al.~\cite{zheng2023gps} achieve generalization to novel humans by incorporating a stereo-depth estimation module, which serves as a partial geometry prior. However, they suffer when given sparse views with few overlappings and thus depth could not be estimated. Therefore, they can only interpolate between very close views. In this work, we aim for a feed-forward generalizable human rendering method that can work when given very sparse inputs with few or no correspondences by leveraging 3D human prior. 

\mypara{Multi-surface representations.}
While utilizing 3D human prior has proven its effectiveness in the human rendering task~\cite{peng2021neural,peng2021animatable,liu2021neural,su2021nerf,habermann2023hdhumans}, representing the geometry gap between human template and the real geometry (\eg loose clothing, hair) is still challenging. 
Some recent literature~\cite{ouyang2022real,kwon2023deliffas,abdal2023gaussian,wang2023adaptive} utilize multi-surface (shell)~\cite{porumbescu2005shell} to represent the geometry displacement. However, the idea from these works is not directly applicable to our generalization task because they either can be only optimized on a single subject or cannot be conditioned on the input subject information (i.e., unconditional generation from noise)~\cite{abdal2023gaussian}. In this paper, we propose multi-scaffold, a multi-surface-based representation that can effectively represent the human geometric details in the \textit{generalization setting}.
\section{Generalizable Human Gaussians (GHG)} \label{sec:method}

\begin{figure}[t]
\centering
\def\arraystretch{0.5}
\includegraphics[width=1\linewidth]{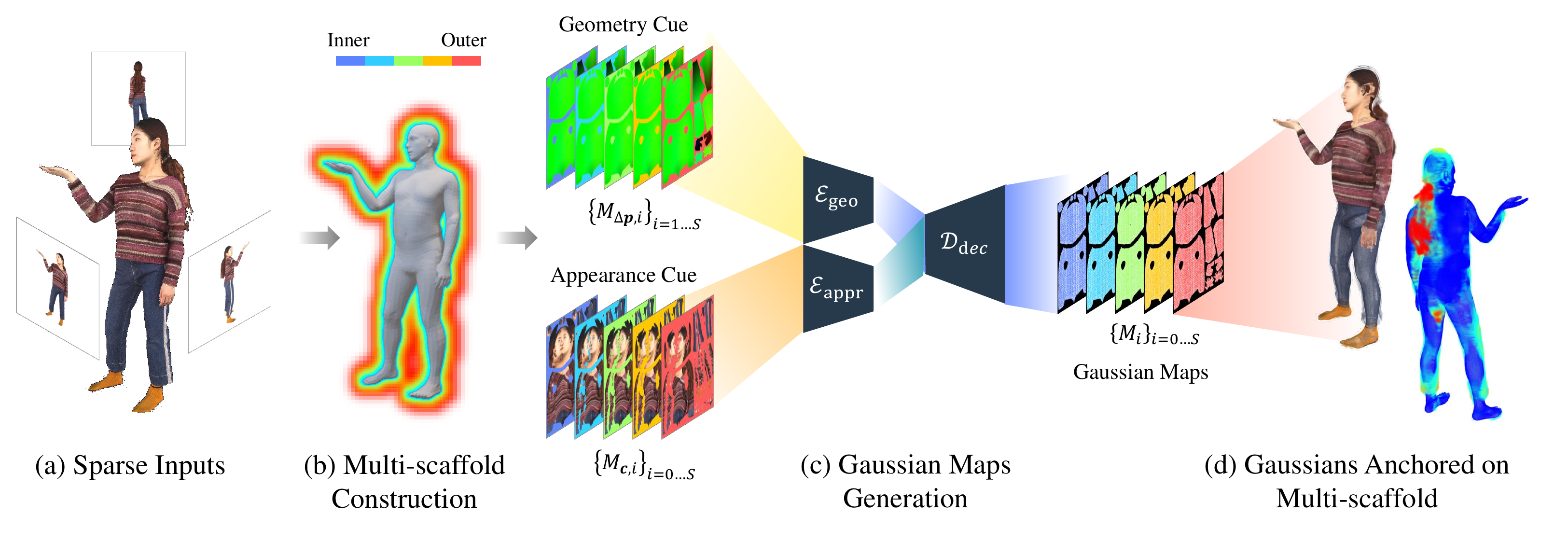}

\caption{\small \textbf{Overview of GHG.} (a) We focus on generalizable human rendering under very sparse view setting. (b) We first construct the multi-scaffolds by dilating the human template surface. The 2D UV space of each scaffold serves to collect the geometry and appearance information from the corresponding 3D locations. (c) The aggregated multi-scaffold input is fed into the network, which generates multi-Gaussian parameter maps. (d) Finally, Gaussians are anchored on the corresponding surface of each scaffold, and rasterized into novel views.
}
\label{fig:overview}
\vspace{-6mm}
\end{figure}

Given a set of multi-view images of a subject (that is not in the training set), along with their camera position and fitted human template (i.e., SMPL~\cite{loper2023smpl}), our goal is to render photo-realistic novel views. 
To address this challenge, we propose Generalizable Human Gaussian (GHG), a feed-forward architecture that does not require any fine-tuning. See Figure~\ref{fig:overview} for an illustration of GHC. 
In this section, we first review the 3D Gaussian Splatting and discuss the motivation of GHG (Section~\ref{sec:backgrounds}).  
In Section~\ref{sec:reformulation}, we introduce the main idea of GHG, which reformulates the Gaussian splat fitting as a regression problem in the 2D UV space of the human template. 
Next, we present our multi-scaffold representation that allows encoding and modeling of the complicated geometric details (Section~\ref{sec:multi_scaffold}).
Finally, we describe the end-to-end training objective in Section~\ref{sec:supervision}.

\subsection{Background and Motivation } \label{sec:backgrounds}
\mypara{Notation.} 
Functions (e.g., neural network mapping) are denoted with uppercase calligraphic letters (e.g., $\mathcal{F}$). Vectors are denoted with bold lowercase letters (e.g., $\bm{p}$). Matrices are denoted with uppercase letters (e.g., $M$). Sets are denoted with bold uppercase letters (e.g., $\bm{\Theta}$).

\mypara{3D Gaussian Splatting (3D-GS).}
The key idea of 3D-GS~\cite{kerbl20233d} is to represent a scene with a set of 3D Gaussians, each of which is characterized by a 3D covariance matrix ${\Sigma}$ and a center (mean) position $\mathbf{p}$:
\vspace{-0.05in}
\begin{equation}
    \label{eq:gaussian}
    \mathbf{G}(\mathbf{x})~= e^{-\frac{1}{2}(\mathbf{x}-\mathbf{p})^{T}{{\Sigma}}^{-1}(\mathbf{x}-\mathbf{p})}.
    \vspace{-0.05in}
\end{equation}
The means of the 3D Gaussians can be initialized by a point cloud using Structure from Motion~\cite{schonberger2016structure} computed from $C$ images. Each Gaussian $\textbf{G}$ is parameterized by $\bm{\Theta} = \{\mathbf{p}, \mathbf{q}, \mathbf{s}, {\alpha}, \boldsymbol{\upeta} \}$ where $\mathbf{p} \in \mathbb{R}^{3}$ is the center position, $\mathbf{q} \in \mathbb{R}^{4}$ is the rotation quaternion,  $\mathbf{s} \in \mathbb{R}^{3}$ is the scaling factor, ${\alpha} \in \mathbb{R}^{1}$ is the opacity, and $\boldsymbol{\upeta} \in \mathbb{R}^{(l+1)^2}$ represents the coefficients of the spherical harmonics (SH) of order $l$.
The covariance matrix is decomposed as ${\Sigma}={R} {S} {S}^{\top} {R}^{\top}$, where ${S}=diag(\mathbf{s}) \in \mathbb{R}^{3 \times 3}$ is the scaling matrix and ${R} \in \mathbb{R}^{3 \times 3}$ is a rotation matrix derived from the quaternion $\mathbf{q}$.

The rendering of a Gaussian set into an image plane is done by approximating the projection of a 3D Gaussian into pixel coordinates along the depth dimension~\cite{kerbl20233d}.
Specifically, for each pixel, the final rendered color $\textbf{c}_\text{pixel}$ is obtained by the $\alpha$-blending of $K$ overlapping Gaussians that are depth-ordered: 
\begin{equation}
    \textbf{c}_\text{pixel} = \sum_{i\in K} \textbf{c}_i \alpha_i \prod_{j=1}^{i-1}(1-\alpha_j),
    \label{eq:render}
\end{equation}
where $\alpha_i$ is the opacity and $\textbf{c}_i$ is the RGB color extracted from SH coefficients. 

\mypara{Motivation.}
Applying GS to our task (i.e., generalizable human rendering from sparse views) is not trivial for the following reasons:
First, the original GS was designed for single-scene optimization, making it difficult to adapt to generalization tasks --- specifically to reconstruct unseen human subjects without model fine-tuning. Second, accurate point cloud initialization requires a substantial number of input images for Structure from Motion. With the number of input images reduced to as few as $3$, vanilla 3D-GS struggles to accurately reconstruct the complex geometry and texture of the human body. See \figref{fig:teaser}-Vanilla 3D-GS as an example of the view-reconstruction achieved with the original GS. Therefore, in this paper, we focus on adapting the 3D Gaussian Splatting for generalizable human rendering from sparse inputs.

\subsection{Learning 3D Gaussians in 2D Human UV Space} \label{sec:reformulation}

\mypara{UV space of human template.}
Our goal is to model a generalizable funtion $\mathcal{F}(\{ {I_c} \}_C) = \{ \bm{\Theta}_n \}_{N_\mathbf{G}}$ that estimates  the parameters of $N_\mathbf{G}$ Gaussians conditioned on the input images $\{ {I_c} \}_C$.
However, due to the complex nature of human geometry that involves articulations and occlusions, it is challenging to regress the parameters only given few sparse observations.
Therefore, we propose to incorporate a 3D geometry prior (i.e., a human template model such as SMPL~\cite{loper2023smpl}) by attaching the Gaussians on the template surface and regressing their parameters in the 2D UV space of the human template. 
Specifically, for every foreground pixel of the UV map, we attach a Gaussian on the corresponding 3D human surface point defined by the UV mapping. 
Then, we regress and store its parameters in the set of 2D UV maps $\bm{M} = \{ M_{\mathbf{p}}, M_{\mathbf{q}}, M_{\mathbf{s}}, M_{\alpha}, M_{\mathbf{c}} \}$. $M_{\mathbf{p}}, M_{\mathbf{q}}, M_{\mathbf{s}}, M_{\alpha}, M_{\mathbf{c}}$ denotes the map for the position, rotation, scaling, opacity, and RGB color, respectively. Each parameter map has the resolution of $H \times W \times D$, where D is the dimension of each parameter.

\mypara{2D CNN-based parameter regression.}
To model the function $\mathcal{F}$, we adopt a 2D  Convolutional Network that provides several benefits. First of all, 2D CNN naturally aggregates the information from neighboring pixels. This helps our system to consider the local context and thus maintain the consistency between the adjacent Gaussian parameters, which both contribute to better reconstruction accuracy. In addition, it facilitates integrating other image-based enhancement models, in our case the inpainting module to hallucinate unobserved regions. In practice, $\mathcal{F}$ is modeled with a U-Net as follows:
\begin{equation} \label{eq:basic}
\mathcal{F} = \mathcal{D}_\text{dec} \bigl( \mathcal{E} ( \{ {I_c} \}_C) \bigl) = \bm{M},
\end{equation}
where $\mathcal{E}$, $\mathcal{D}_\text{dec}$ is the U-Net-based encoder and decoder, respectively. $\{ {I_c} \}_C$ denotes the input images, and $M$ is the set of Gaussian parameter maps.

\mypara{Reformulation.}
In our formulation, $\mathcal{F}$ regresses the set of 2D parameter maps $\bm{M} = \{ M_{\mathbf{p}}, M_{\mathbf{q}}, M_{\mathbf{s}}, M_{\alpha}, M_{\mathbf{c}} \}$.
Since the Gaussian positions are fixed on the human template surface, the position map $M_{\mathbf{p}}$ is computed by rasterizing the vertex position of the human template on the 2D UV space.
The RGB map $M_{\mathbf{c}}$ is computed as the weighted average of corresponding pixels from all observed views. Specifically, for each pixel in $M_{\mathbf{c}}$, we find their projections to all visible source images and average the source RGB values weighted by visibility:
\vspace{-0.05in}
\begin{equation} \label{eq:basic_rgb_map}
M_{\mathbf{c}} = \sum_{c=1}^{C} {\text{W}_\text{c}(\bm{P})} \cdot {\mathrm\Pi} \bigl( {I}_{c}, \text{Proj}_\text{c}(\bm{P}) \bigl). 
\vspace{-0.05in}
\end{equation}
$\bm{P}$ is the Gaussian center positions (i.e., foreground pixel values of the position map $M_p$). $\text{W}_\text{c}(\bm{P})$ is the normalized visibility of 3D positions $\bm{P}$ for the $c$-th camera. $C$ is the number of total input views. $I_c$ is the $c$-th input view image. $\mathrm{\Pi}$ denotes the bilinear sampling operator. $\text{Proj}_\text{c}$ denotes the 3D to 2D projection with respect to the $c$-th camera. $ {\mathrm\Pi} \bigl( {I}_{c}, \text{Proj}_\text{c}(\bm{P}) \bigl)$ returns an image that is the result of the of interpolating ${I}_{c}$ in the projected coordinates of $\bm{P}$.

Since $M_{\mathbf{p}}$ and $M_{\mathbf{c}}$ are already computed, the regression of the parameter map $\bm{M}$ is reduced to $\bm{M}=\{M_{\mathbf{q}}, M_{\mathbf{s}}, M_{\alpha}\}$.
To effectively regress the Gaussian parameter maps, we provide $\mathcal{F}$ with the geometry and appearance cue which have complementary attributes.

The appearance cue provides information of geometric details and how they should look like. The geometry cue facilitates the optimization of the Gaussian parameters to match the appearance details.
To provide the geometry cue, we encode the position map $M_{\mathbf{p}}$ using the geometry encoder $\mathcal{E}_\text{geo}$. The appearance cue is obtained by encoding the RGB map $M_{\mathbf{c}}$ using the appearance encoder $\mathcal{E}_\text{appr}$. 
Now, we adapt the Equation~\ref{eq:basic} to condition the parameter map generation on the geometry and appearance cue, as:

\begin{equation} \label{eq:reformulation}
\mathcal{D}_\text{dec} \bigl( \mathcal{E}_\text{geo}(M_{\mathbf{p}}),\mathcal{E}_\text{appr}(M_{\mathbf{c}}) \bigl) = \bm{M}.
\end{equation}
\mypara{Inpainting.}
It is inevitable to have unobserved regions under very sparse view settings. This results in blurriness or missing texture in some areas. To address this issue we incorporate into our architecture a 2D inpainting method. In particular, we create a set of pseudo ground truth texture maps by transferring the ground truth mesh texture map into the human template UV space (see Appx-Fig.10). On this dataset, we train an attention-based generative model $\mathcal{G}_\text{inpaint}$~\cite{yu2018free,yu2018generative} to inpaint the missing regions present in the human template UV space RGB map. At the inference time, we inpaint the RGB map $M_{\mathbf{c}}$ with $\mathcal{G}_\text{inpaint}$. We would like to note that this is possible because our 2D CNN-based system facilitates the combination with a 2D-based inpainting module. Our parameter map regression is again adapted to:
\begin{equation} \label{eq:reformulation_inpaint}
\mathcal{D}_\text{dec} \Bigl( \mathcal{E}_\text{geo}(M_{\mathbf{p}}),\mathcal{E}_\text{appr} \bigl( \mathcal{G}_\text{inpaint}(M_{\mathbf{c}}) \bigl) \Bigl) = \bm{M}.
\end{equation}
Please refer to the supplementary materials for details of the inpainting network.

\begin{figure}[t]
\centering
\def\arraystretch{0.5}
\includegraphics[width=0.90\linewidth]{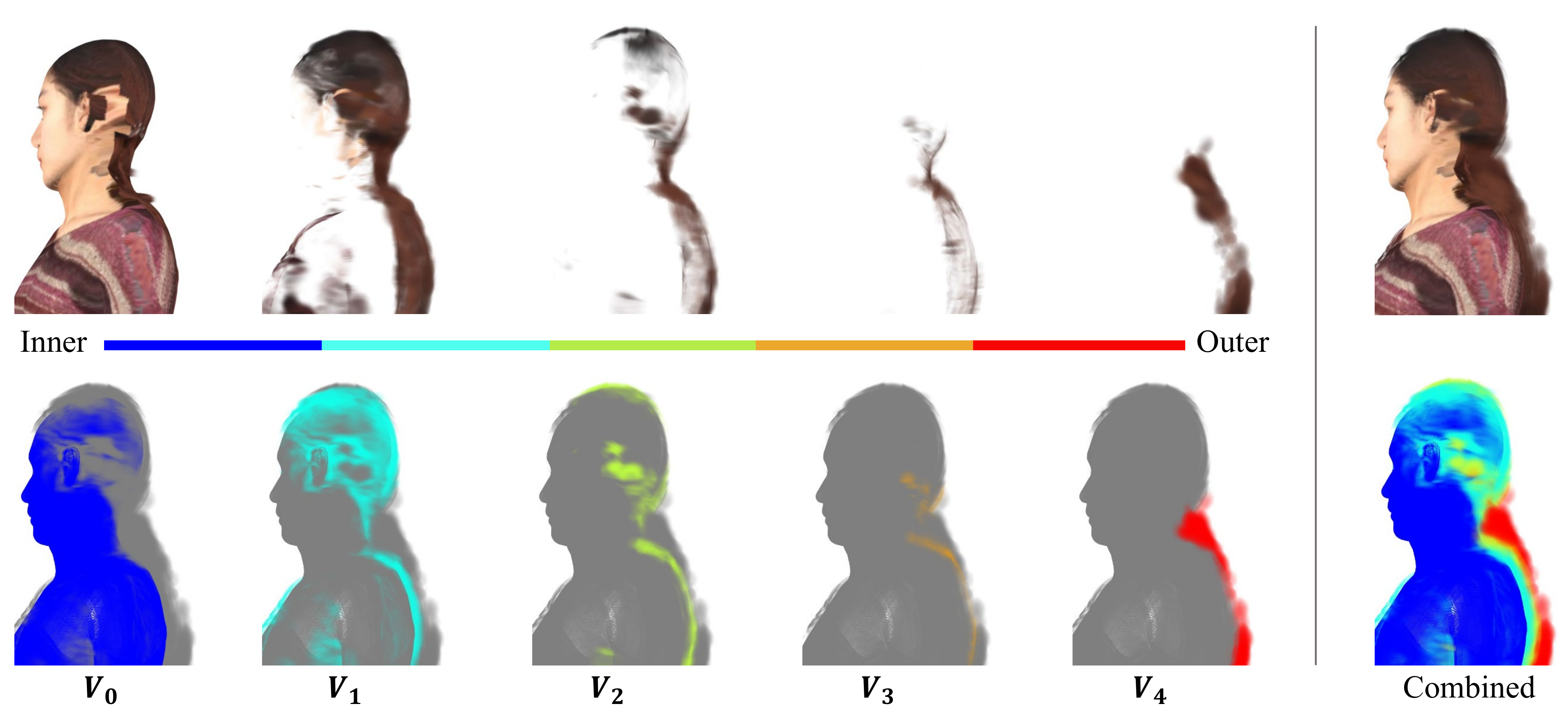}
\vspace{-3mm}
\caption{\small \textbf{Illustration of multi-scaffold representation.}
Each column shows different scaffold levels, with the last column illustrating their combined effect. The top part shows the RGB representation, while the bottom part highlights affected regions, with grey indicating unaffected areas.
}
\label{fig:multi}
\vspace{-3mm}
\end{figure}

\subsection{Modeling Geometric Details with Multi-scaffolds} \label{sec:multi_scaffold}

The utilization of a human template model helps to reconstruct the shape and appearance with sparse views. However, this is not enough to effectively represent accurately details that are offseted from the human surface such as hair or loose clothing due to the following reasons:
%
(1) The appearance details deviating from the template surface (e.g., ponytail) cannot be accurately represented with the input appearance cue (i.e., RGB map $M_c$ in \eqnref{eq:reformulation_inpaint}). 

Therefore, to narrow this geometry gap, we propose to utilize multiple scaffolds constructed through dilation of the human template mesh. These multi-scaffold representation facilitates the effective encoding of the geometry gap information into the input, and allows for more versatile output Gaussians to represent the displacement details more accurately.
Specifically, we create the multiple scaffolds $\{ \bm{V}_i \}_{i= 1 \ldots S}$ by offsetting the human template vertices $V_0=\{ \mathbf{v}_\text{0,j} \}$ along its outward vertex normal direction:
%
\begin{equation}  \label{eq:multi_scaffold}
    \bm{V}_i = \{ \mathbf{v}_\text{i,j} |  \mathbf{v}_\text{i,j} = \mathbf{v}_\text{0,j} + i \cdot d \cdot \mathbf{\hat n}_\text{j} \},
\end{equation}
%
where $\mathbf{v}_{i,j}$ is the $j$-th vertex of $i$-th outer scaffold, $\mathbf{\hat n}_\text{j}$ is the $j$-th vertex normal, $d$ defines the offset between scaffolds, $S$ is the number of outer scaffolds. We use ${d=1 cm}$ and $S=4$ in the experiments.

\mypara{Input.}
We adapt the geometry and appearance cues to aggregate information from the entire level of scaffolds.
We redefine the geometry cue as the feature extracted from the concatenation of offset maps which record the displacement between each scaffold:
\begin{equation}  \label{eq:multi_scaffold_geo}
    \mathcal{E}_\text{geo}( M_{\Delta \mathbf{p},1}  \oplus \ldots \oplus M_{\Delta \mathbf{p},S} ), \text{ where} ~ M_{\Delta \mathbf{p},i}= M_{\mathbf{p},i} - M_{\mathbf{p},i-1}.
\end{equation}
$M_{\Delta \mathbf{p},i}$, $M_{\mathbf{p},i}$ is the offset map and position map of the $i$-th scaffold, respectively. $\oplus$ is the concatenation operation. The appearance cue is redefined as: 
%
%
\begin{equation}  \label{eq:multi_scaffold_appr}
    \mathcal{E}_\text{appr}( M_{\mathbf{c},0} \oplus \ldots \oplus M_{\mathbf{c},S} ), \text{ where} ~ M_{\mathbf{c},i} = \sum_{c=1}^{C} {\text{W}_\text{c} (\bm{P}_i)} \cdot \mathrm{\Pi}({I}_\text{c}, \text{Proj}_\text{c}(\bm{P}_i)).
\end{equation}
Here $M_{\mathbf{c},i}$, $\bm{P}_i$ is the RGB map and the Gaussian center positions corresponding to the $i$-th level scaffold, respectively. Note that inpainting is done only to the RGB map of $0$-th level scaffold (i.e., $M_{\mathbf{c},0} = \mathcal{G}_\text{inpaint}(M_{\mathbf{c},0})$)

\mypara{Output.}
There are numerous possible design choices to model the displacement details such as hair. For example, the scaling can be enlarged to cover the gap, or we could directly learn Gaussian mean offsets. However, we empirically found out that these lead to unstable training and hinder the system from converging because of the high degree-of-freedom (see \figref{fig:ablation}). Therefore, we attach Gaussians on each scaffold, and regress their parameters \textit{within each scaffold}. This is realized by confining the maximum scaling of the Gaussians as the offset between scaffolds. 
Our final formulation is defined as:
\begin{equation} \label{eq:reformulation_final}
\mathcal{D}_\text{dec} \bigl( \mathcal{E}_\text{geo}( M_{\Delta \mathbf{p},1}  \oplus \ldots \oplus M_{\Delta \mathbf{p},S} ),\mathcal{E}_\text{appr}( M_{\mathbf{c},0} \oplus \ldots \oplus M_{\mathbf{c},S} ) \bigl) = \{ \bm{M}_i \}_{i=0 \dots S}.
\end{equation}
where $\{ \bm{M}_i \}$ is the set of parameter maps corresponding to the $i$-th level scaffold.
\subsection{Training and Optimization} \label{sec:supervision}

\mypara{Gaussian parameter map regressor.}
To train our Gaussian parameter map regressor $\mathcal{F}$ (i.e., $\mathcal{E}_\text{geo}$, $\mathcal{E}_\text{appr}$, $\mathcal{D}_\text{dec}$), we employ multi-view RGB and mask supervision. Specifically, we sample $N$ target views from the positions in between the input views and generate the RGB and mask predictions. The predictions are supervised by minimizing the loss objective 
$\mathcal{L} = \frac{1}{N}(\lambda_{1}\cdot\mathcal{L}_{1} + \lambda_\mathrm{ssim}\cdot\mathcal{L}_\mathrm{ssim}+ \lambda_\mathrm{mask}\cdot\mathcal{L}_\mathrm{mask}$),
where $\mathcal{L}_{1}$, $\mathcal{L}_\mathrm{ssim}$ are $\text{L}_{1}$ and SSIM loss~\cite{wang2004ssim} computed between the ground truth and predicted RGB images, respectively. $\mathcal{L}_\mathrm{mask}$ is the Binary Cross Entropy loss computed between the ground truth and predicted foreground mask. We use $N=3$, $\lambda_{1}=0.8$, $\lambda_\mathrm{ssim}=0.2$, $\lambda_\mathrm{mask}=0.02$ in our experiments. 
We used a single GPU with 20G memory during training. AdamW optimizer~\cite{adamw} with an initial learning rate of $2e^{-4}$ was used. We train the parameter regressor for $100k$ iterations with a single batch size, which takes around 10 hours.

\mypara{Inpainting network.}
When training the inpainting network $\mathcal{G}_\text{inpaint}$, $\mathcal{L}_{G}=\lambda_\text{rec}\cdot \mathcal{L}_\text{rec}+\lambda_\text{adv}\cdot \mathcal{L}_\text{adv}$ is minimized, where $\mathcal{L}_\text{rec}$, $\mathcal{L}_\text{adv}$ are $\text{L}_1$ loss and adversarial loss computed between the inpainted results and pseudo ground truth. $\lambda_\text{rec}=10$ and $\lambda_\text{adv}=1$ are used in training.
The loss objective for the inpainting discriminator $\mathcal{D}_\text{inpaint}$ is defined the discriminator loss between the inpainted image and pseudo ground truth.
$\mathcal{G}_\text{inpaint}$ and $\mathcal{D}_\text{inpaint}$ are trained alternatively for $40$ epochs. We use Adam optimizer~\cite{kingma2014adam} with an initial learning rate of $1e^{-4}$ and decay the learning rate by half every $10$ epoch. It takes around $4$ hours to train the inpainting module on a single GPU with a batch size of $1$. Note that $\mathcal{G}_\text{inpaint}$ is trained only once separately from the Gaussian map regressor, and it is a general model that works for different new subjects at inference.
\begin{table}[t]
\centering
\small
\tabcolsep 6pt
\caption{\small{\textbf{Comparison with NeRF-based methods for (a) in-domain and (b) cross-domain sparse view synthesis.} For all the methods, we use \textbf{$3$} views during both training and testing. GHG achieves competitive results for both settings. TH: THuman~\cite{thuman}. RP: RenderPeople~\cite{renderpeople}.} See Figure~\ref{fig:comparison} and \ref{fig:cross_dataset} for qualitative results.}
\vspace{-2mm}
\begin{tabular}{lcccccc}
 & \multicolumn{3}{c}{\textbf{(a) In-domain: TH $\rightarrow$ TH}} & \multicolumn{3}{c}{\textbf{(b) Cross-domain: TH $\rightarrow$ RP}} \\
 \cmidrule(lr){2-4} \cmidrule(lr){5-7}
 Method & \multicolumn{1}{c}{PSNR$\uparrow$} & \multicolumn{1}{c}{LPIPS$\downarrow$} & \multicolumn{1}{c}{FID$\downarrow$} & \multicolumn{1}{c}{PSNR$\uparrow$} & \multicolumn{1}{c}{LPIPS$\downarrow$} & \multicolumn{1}{c}{FID$\downarrow$} \\
 \toprule
 NHP~\cite{kwon2021neural} & \textbf{23.32} & 184.69 & 136.56 & 22.34 & 172.56 & 137.23 \\
NIA~\cite{kwon2023neural} & 23.20 & 181.82 & 127.30 & \textbf{22.45} & 168.15 & 124.80 \\
\textbf{GHG (ours)} & 21.90 & \textbf{133.41} & \textbf{61.67} & 21.02 & \textbf{137.73} & \textbf{60.85} \\
\bottomrule
\end{tabular}
\label{tab:comparison_nerf}
\vspace{-4mm}
\end{table}

\section{Experiments} \label{sec:experiments}
\subsection{Baselines, Datasets, and Metrics}

\mypara{Baselines.}
We benchmark our method against state-of-the-art generalizable human rendering techniques from two categories: 3D human template-conditioned NeRF methods NHP~\cite{kwon2021neural} and NIA~\cite{kwon2023neural}, and depth-based 3D Gaussian method GPS-Gaussian~\cite{zheng2023gps}. Additionally, we compared with the original vanilla 3D Gaussians~\cite{kerbl20233d}, which are optimized per subject. 

\mypara{Datasets.}
We conducted experiments on two datasets: the THuman ~\cite{thuman} and RenderPeople~\cite{renderpeople} dataset. 
The THuman dataset comprises 526 high-quality 3D scans, texture maps, and corresponding SMPL-X parameters. 100 subjects were reserved for the evaluation, following GPS-Gaussian~\cite{zheng2023gps}.
The RenderPeople dataset encompasses 3D human scans representing diverse clothing styles, races, and ages, totaling 956 subjects split into 756 train and 200 test subjects. SMPL-X parameters were estimated using off-the-shelf methods~\cite{bhatnagar2020ipnet,bhatnagar2020loopreg}.

\begin{figure}[ht]
\centering
\vspace{-9mm}
\includegraphics[width=1\linewidth]{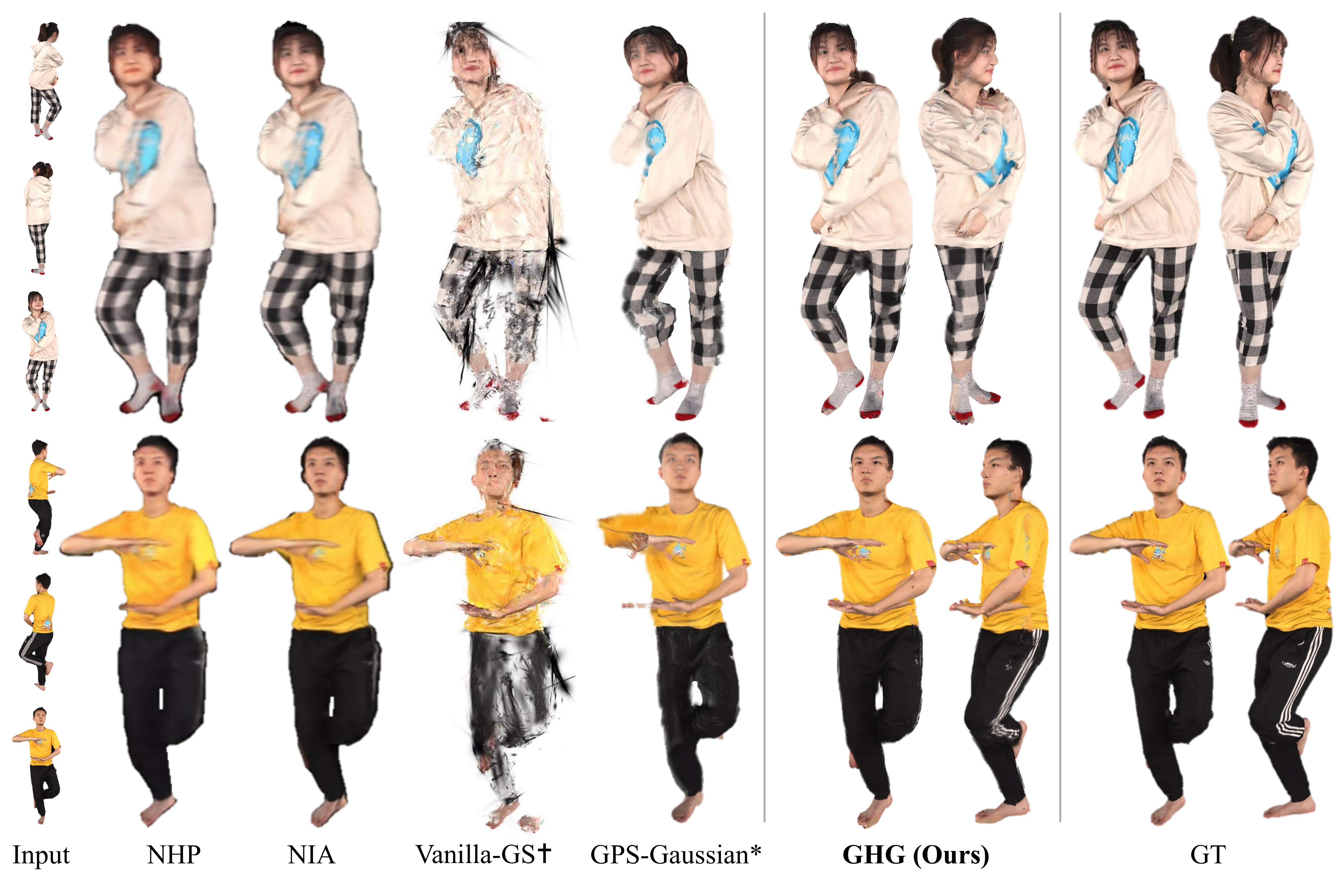}
\vspace{-5mm}
\caption{\small \textbf{Qualitative comparisons.}
 All methods are trained and tested on THuman dataset~\cite{thuman}. $\dagger$Unlike the other methods, Vanilla-GS~\cite{kerbl20233d} is per-subject optimized on the testing subjects. *GPS-Gaussian~\cite{zheng2023gps} is trained and tested with 5 input views, whereas NHP~\cite{kwon2021neural}, NIA~\cite{kwon2023neural} and our method are trained and tested with 3 input views.
}
\label{fig:comparison}
\end{figure}

\mypara{Metrics.} 
We generated images at a resolution of $1024 \times 1024$ for evaluating our results. To assess the quality of our results, we employed several metrics. Initially, we utilized the peak signal-to-noise ratio (PSNR), a standard metric. However, PSNR may not fully reflect human perception, as it can assign a low error to very blurry and unrealistic results~\cite{zhang2018unreasonable}. Therefore, we also incorporated the learned perceptual image patch similarity (LPIPS)\cite{zhang2018unreasonable} and the Fréchet inception distance (FID)\cite{heusel2017gans}, which better align with human perceptions.

\subsection{Comparison with NeRF-based methods}

We compare with NHP~\cite{kwon2021neural} and NIA~\cite{kwon2023neural}, which are the two competitors that are most similar to our settings in that (1) they focus on generalizable human rendering from very sparse (i.e., $3$) input views and (2) use the human template as 3D prior. Ours, NHP, and NIA are trained / evaluated on the THuman dataset with the same NHP protocol, where three randomly chosen input views are used during training and the same three canonical views are used during evaluation.

\mypara{In-domain generalization.}
Table~\ref{tab:comparison_nerf}-(a) shows the in-domain generalization result where we evaluate on test subjects from THuman dataset. We achieve the best performance on the perception-based metrics LPIPS and FID, and comparable PSNR. As shown in Figure~\ref{fig:comparison}, the single-layer representation of NHP and NIA where the features are aggregated on a single surface of a naked body leads to the mixture of visual details and produce blurry results. On the other hand, our method collects visual information from the multi-scaffold and thus recovers sharp and high-frequency details including hair, wrinkles, and logos. 

\begin{figure}[t]
\centering
\includegraphics[width=0.9\linewidth]{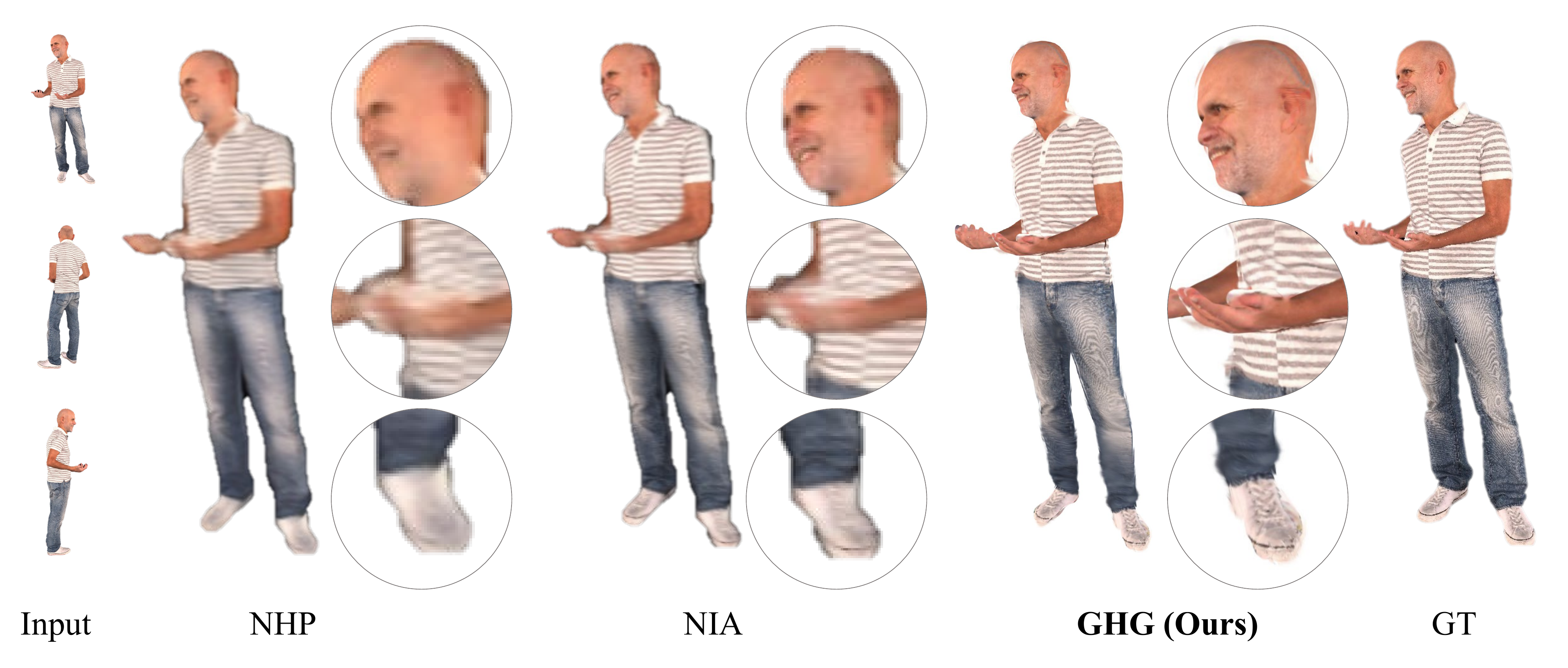}
\vspace{-4mm}
\caption{\small \textbf{Qualitative results on cross-domain generalization.} We train the models on THuman dataset~\cite{thuman} and test on Renderpeople dataset~\cite{renderpeople} without model finetuning. GHG can render high-frequent details and accurate geometry of the novel subject.
}\vspace{-7mm}
\label{fig:cross_dataset}
\end{figure}

\mypara{Cross-domain generalization.}
To confirm the cross-dataset generalizability of our approach, we train a model on the THuman dataset and evaluate it on the challenging RenderPeople dataset without any test-time optimization. The RenderPeople dataset exhibits a more diverse data distribution compared to the training dataset (THuman), encompassing variations in race, age, and apparel. Yet, our GHG significantly outperforms NHP and NIA on the perception-based metrics LPIPS and FID. In Figure~\ref{fig:cross_dataset}, our method recovers fine details such as facial expressions, clothing patterns, and textures. However, since PSNR is pixel-wise computed, a slight deviation from the ground truth can lead to a lower score. This can explain our lower performance in terms of PSNR in Table~\ref{tab:comparison_nerf}-(a),(b), while NHP with blurry results achieves the highest performance.

\subsection{Comparison with Gaussian Splatting-based methods}
\begin{table}[t]
\centering
\small
\tabcolsep 12pt
\caption{\small{\textbf{Comparison with Gaussian Splatting-based methods on the THuman dataset\cite{thuman}.} Due to GPS-Gaussian~\cite{zheng2023gps} requiring at least 5 input views for reasonable results, we train and test all methods with 5 views.}} 
\vspace{-2mm}
\begin{tabular}{lccc}
 Method & \multicolumn{1}{c}{PSNR$\uparrow$} & \multicolumn{1}{c}{LPIPS$\downarrow$ } & \multicolumn{1}{c}{FID$\downarrow$ } \\
 \toprule
Vanilla-GS (per-subject)~\cite{kerbl20233d} & 17.62 & 220.30 & 210.03\\
GPS-Gaussian~\cite{zheng2023gps} & 20.69 & \textbf{123.30} & 46.26  \\
\textbf{GHG (ours)} & \textbf{22.06}  & 132.42  & \textbf{37.10} \\
\bottomrule
\end{tabular}\label{tab:comparison_gs}
\vspace{-3mm}
\end{table}

We show the comparison with GPS-Gaussian~\cite{zheng2023gps} and the original vanilla 3D Gaussians~\cite{kerbl20233d}.
Although the original GPS-Gaussian does not focus on sparse view synthesis as ours, we include comparison with them because we are both 3D Gaussian-based methods and explore generalization onto unseen human subjects.
In our exploration of GPS Gaussian, we observed that GPS-Gaussian requires a substantial overlap between inputs for stereo-depth computation and cannot perform adequately with fewer than five views. Therefore, for comparison purposes in Table~\ref{tab:comparison_gs}, we employ five uniformly distributed input views for training and evaluation of GPS-Gaussian, vanilla 3D Gaussian, and our method. Unlike the other methods, vanilla Gaussian is per-subject optimized and thus trained on the testing human subjects.
As presented in Table~\ref{tab:comparison_gs}, our approach achieves comparable results to GPS-Gaussian~\cite{zheng2023gps} with better PSNR and FID scores, while significantly outperforms the vanilla Gaussian method. Visual comparisons in Figure~\ref{fig:comparison} reveal that our model, trained and tested with 3 input views, exhibits more accurate geometries and finer detail reconstruction compared to GPS-Gaussian, trained and tested with 5 input views. Particularly, GPS-Gaussian suffers from inaccurate geometry and missing contents, possibly due to its inherent high demand for larger overlap between input views.

\subsection{Ablation Studies and Analyses}
\begin{table}[t]
\centering
\small
\tabcolsep 6pt
\caption{
\small{
\textbf{Ablation study on the multi-scaffold representation.} \textbf{S:} single scaffold. \textbf{S$\star$:} single scaffold with a large scale. \textbf{S$\dagger$:} single scaffold with a learnable offset. \textbf{\cmark:} multiple scaffolds.
}}\label{tab:abl_multi_scaffold}
\vspace{-3mm}
\begin{tabular}{ccccccc}
\multicolumn{1}{l}{} & \multicolumn{2}{c}{Input Scaffold} & Output Scaffold &  &  &  \\
\cmidrule{2-4}
\multicolumn{1}{l}{} & Geometry & Appearance & Gaussian Map & \multicolumn{1}{c}{PSNR$\uparrow$} & \multicolumn{1}{c}{LPIPS$\downarrow$} & \multicolumn{1}{c}{FID$\downarrow$} \\
\toprule
a & S & S & S  
    & 22.30     & 145.74    & 84.38  \\
b & S & S & \cmark  
    & 22.44     & 142.84    & 73.91  \\
c & S & \cmark  & \cmark   
    & 22.96     & 136.81    & 72.43  \\
d & \cmark  & S & \cmark  
    & 22.60     & 144.27    & 81.31  \\ 

\midrule

e & \cmark  & \cmark & ~S$\star$
    & 23.11     & 145.03    & 90.16  \\ 
f  & \cmark  & \cmark & ~S$\dagger$  
    & \bf{23.39}     & 145.55    & 87.49  \\ 
\midrule
g  & \cmark  & \cmark  & \cmark  
    & {21.90}     & \bf{133.41}    & \bf{61.67} \\
\bottomrule
\end{tabular}
\end{table}

\begin{figure}[t]
\centering
\def\arraystretch{0.5}
\includegraphics[width=0.9\linewidth]{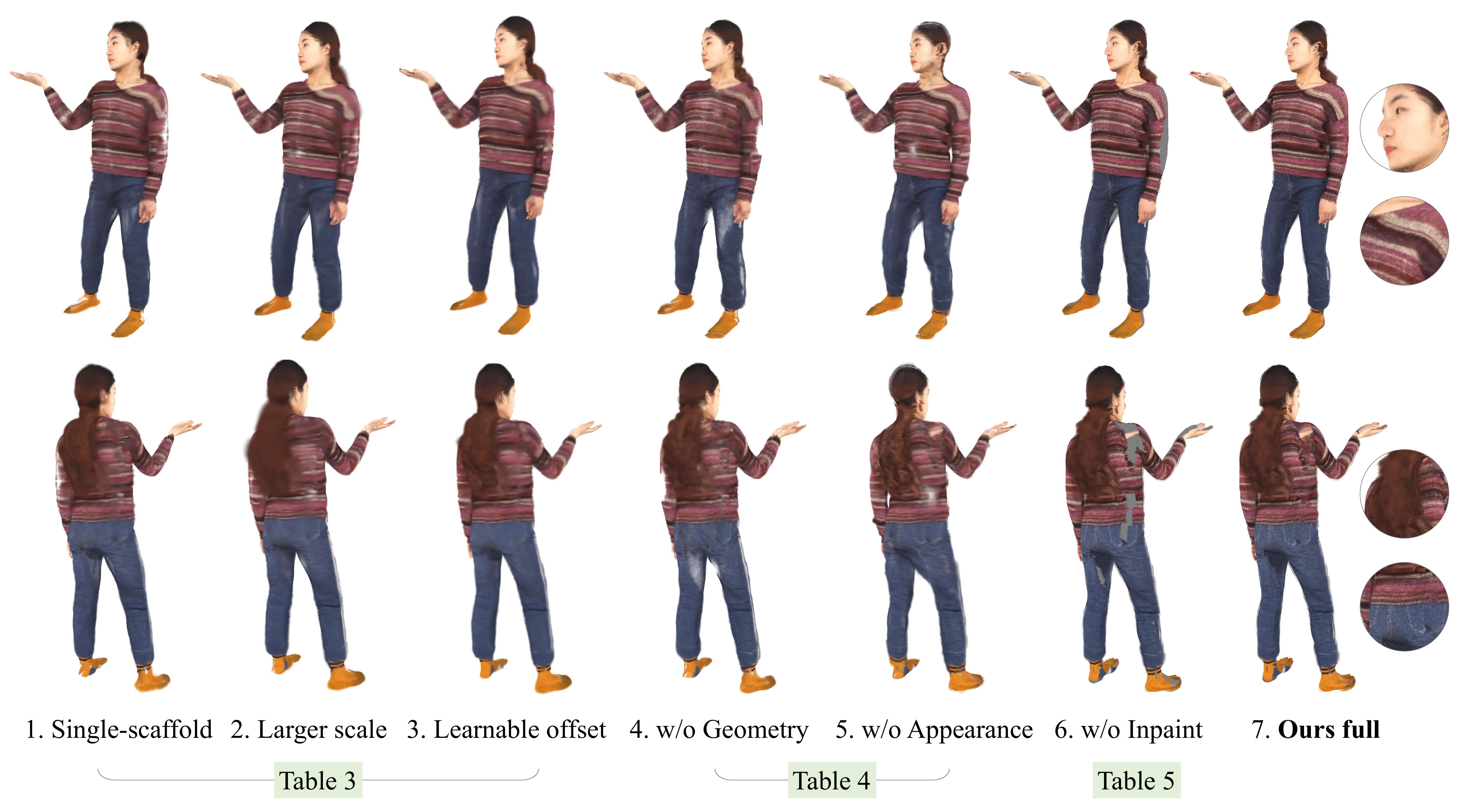} 
\caption{\small \textbf{Ablation studies.} 1) Result only using the template mesh. 2) Illustrates the result of using a larger scale. 3) Depicts the result of learning the Gaussian offset. 4) Shows the model devoid of geometry information. 5) Illustrates the model without appearance cues. 6) Shows the model without inpainting. 7) Presents our model.
}
\label{fig:ablation}
\vspace{-2mm}
\end{figure}

We conducted ablation studies on the THuman dataset, evaluating variants of our GHG model on unseen subjects in Figure~\ref{fig:ablation} and Table~\ref{tab:abl_multi_scaffold}, \ref{tab:abl_input}, and \ref{tab:abl_inpainting}. 

\mypara{Effect of multi-scaffold representation.} 
We examine the impact of the multi-scaffold representation for different configurations of inputs/outputs, see Table~\ref{tab:abl_multi_scaffold}.
First, we trained three input variants where the multi-scaffold input is either partially (c, d) or not used at all (b). We retained the multi-scaffold output (i.e., multi-Gaussian map generation). The lack of multi-layer geometry and appearance information leads to perceptual performance degradation.
Second, we designed two output variants where only a single Gaussian map is generated, where we maintained the multi-scaffold input. In our original model, scale of each Gaussian map is confined up to the distance between its next scaffold (i.e., $1cm$). To represent the displacement between the template model and the real geometry, either scale is allowed to grow up to $4cm$ (Table~\ref{tab:abl_multi_scaffold}-(e)) or scale is still confined to $1cm$ but we additionally learn a Gaussian center offset that can move up to $4cm$ (Table~\ref{tab:abl_multi_scaffold}-(f)). However, as shown in Figure~\ref{fig:ablation}-(2),(3), the lack of regularization generates blurry results. The lowest performance of output variants without multi-scaffold representation in Table~\ref{tab:abl_multi_scaffold}-(e),(f) again validates our design choice where we build 3D Gaussians on multiple scaffolds.

\mypara{Importance of geometry and appearance cue.}
To study the effect of integration of the geometry and appearance cue, we train a variant with the geometry cue completely removed (Table~\ref{tab:abl_input}-first row, Figure~\ref{fig:ablation}-4) and a variant with appearance cue removed (Table~\ref{tab:abl_input}-second row, Figure~\ref{fig:ablation}-5). Removing one of them leads to visual artifacts and failure of recovering geometry gap such as hair.

\mypara{Effect of inpainting.}
Under our sparse-view setting, it is inevitable to have insufficient observations (Figure~\ref{fig:ablation}-6). Our 2D architecture allows us to easily combine with the 2D-based inpainting module and hallucinate the unobserved regions, thus leads to improved quality (Figure~\ref{fig:ablation}-7, Table~\ref{tab:abl_inpainting}).

\begin{table}[t]
    \begin{minipage}{.5\linewidth}
      \centering
        \small
        \tabcolsep 4.5pt
        \caption{
        \small{
        \textbf{Ablation study on the geometry cue and appearance cue.} \xmark/\cmark indicates completely remove/keep the encoding branch. See Figure~\ref{fig:overview} for an illustration.
        }}\label{tab:abl_input}
        \vspace{-2mm}
        \begin{tabular}{ccccc}
        Geo. & App. & PSNR$\uparrow$    & LPIPS$\downarrow$ & FID$\downarrow$    \\
        \toprule
        \xmark & \cmark & \bf{22.74}     & 138.35    & 69.12 \\
        \cmark & \xmark  & 21.83     & 146.05    & 78.56 \\
        \cmark & \cmark & {21.90}     & \bf{133.41}    & \bf{61.67} \\
        \bottomrule
        \end{tabular}
    \end{minipage}
    \begin{minipage}{.48\linewidth}
            \centering
            \small
            \tabcolsep 4.5pt
            \caption{
            \small{
            \textbf{Ablation study on the texture inpainting network.} \xmark/\cmark indicates without/with the inpainting network on the 2D UV space, respectively. Please see Figure~\ref{fig:ablation} for a comparison result.
            }}\label{tab:abl_inpainting}
            \vspace{-2mm}
            \begin{tabular}{cccc}
            Inpainting   & PSNR$\uparrow$    & LPIPS$\downarrow$ & FID$\downarrow$  \\
            \toprule
            \xmark & 21.65     & 135.42    & 67.15 \\
            \cmark & \bf{21.90}     & \bf{133.41}    & \bf{61.67} \\
            \bottomrule
            \end{tabular}
    \end{minipage} 
\end{table}

\vspace{-0.1in}
\section{Conclusion}
We present Generalizable Human Gaussians (GHG), a feed-forward architecture capable of synthesizing novel views of new humans using sparse input views, without the need for test-time optimization. Our key insight is the reformulation of 3D Gaussian parameter optimization into the generation of parameter maps within the 2D human UV space. This allows us to leverage the human geometry prior, addressing challenges such as articulations and self-occlusions. Additionally, by framing the task as a 2D problem, we can exploit local neighboring information and integrate 2D-based inpainting modules to hallucinate unobserved regions. Finally, we propose a multi-scaffold approach to effectively represent and bridge the geometry gap between the human template and real human geometry. Experimental results show that our method can generate high-quality renderings surpassing state-of-the-art approaches.

\small{
\bibliographystyle{splncs04}
\bibliography{main}}

\clearpage
\appendix
\section{Appendix - Overview}
This appendix is organized as follows: Sec.~\ref{sec:limitations} discusses the limitations and future works; Sec.~\ref{sec:societal_impacts} presents the societal impacts our work can have; Sec.~\ref{sec:additional_results} shows additional results including video results, comparison with single-view methods, ablations on number of outer scaffolds, ablation study with different loss supervision, ablations on the number of input views at inference time, and runtime at inference.
Sec.~\ref{sec:reproducibility} provides information regarding reproducibility, which includes implementation details.

\section{Limitations and Future Works} \label{sec:limitations}

Although our method achieves state-of-the-art results in terms of visual quality and runtime, it is not free from limitations.
(1) While our method effectively compensates for minor inaccuracies in SMPL-X estimations through the use of multi-scaffolds, significant deviations in SMPL-X from the input images could compromise the quality of our results, as our Gaussians are anchored to the SMPL-X surface. 
(2) Currently, the number of scaffolds is determined empirically. It would be an interesting direction to explore adaptive scaffolds based on subject attributes (e.g., loose or tight clothing).
(3) The performance of our inpainting network is constrained by the small number of ground truth texture maps available during training, which in turn limits its ability to generate detailed hallucinations when given a single-view input. Therefore, integrating and fine-tuning generative models trained on extensive datasets (e.g., Stable Diffusion model~\cite{rombach2022high}) could substantially improve our network's hallucination capabilities and generalizability, which is a promising direction for future work.

\section{Societal Impacts} \label{sec:societal_impacts}
Our proposed method can push immersive entertainment and communication to a more affordable setting. For example, our work has the potential to enhance the accessibility of telepresence experiences by facilitating the creation of avatars from minimal RGB images. Moreover, the technology presents benefits to film and game production by enabling efficient synthesis of large-scale 3D human avatars with low costs.

However, our work might also introduce potential challenges, primarily related to the accessible creation of realistic human images. This could lead to deep-fake human avatars on social media, with implications for misinformation and the degradation of trust in digital content. To mitigate such risks, it is urgent to promote ethical guidelines and regulations on synthetic media. We strongly appeal transparent use of such technology as it should align with societal interests and foster trust rather than skepticism.

\section{Additional results} \label{sec:additional_results}

\subsection{Video results}
Video results of comparison with the state-of-the-art baselines on the in-domain generalization task (i.e., trained and tested on THuman 2.0 dataset~\cite{thuman}) and cross-dataset generalization task (i.e., trained on THuman 2.0 and tested on RenderPeople~\cite{renderpeople}) can be found in the project website\footnote{https://humansensinglab.github.io/Generalizable-Human-Gaussians}. For the in-domain generalization task, we compare our GHG with (1) human template-conditioned NeRF, generalization from sparse view methods NHP~\cite{kwon2021neural} and NIA~\cite{kwon2023neural}, and (2) generalizable 3D Gaussian Splatting for human rendering method GPS-Gaussian~\cite{zheng2023gps}. Note that GPS-Gaussian is trained and tested with 5 input views due to the rectification requirement. NHP, NIA, and ours are trained and tested with 3 input views. For the cross-dataset generalization task, we show comparison with our main baselines NHP and NIA. Our method can recover sharp and fine details compared to human template-conditioned NeRF baselines. Due to the lack of full 3D prior, GPS-Gaussian suffers in maintaining multi-view consistency between the novel views generated using different input views. On the other hand, ours maintains robust and accurate geometry reconstruction utilizing the 3D human template.

\subsection{Comparison with single-view methods}
\begin{figure}[t]
  \centering
  \includegraphics[width=0.98\linewidth]{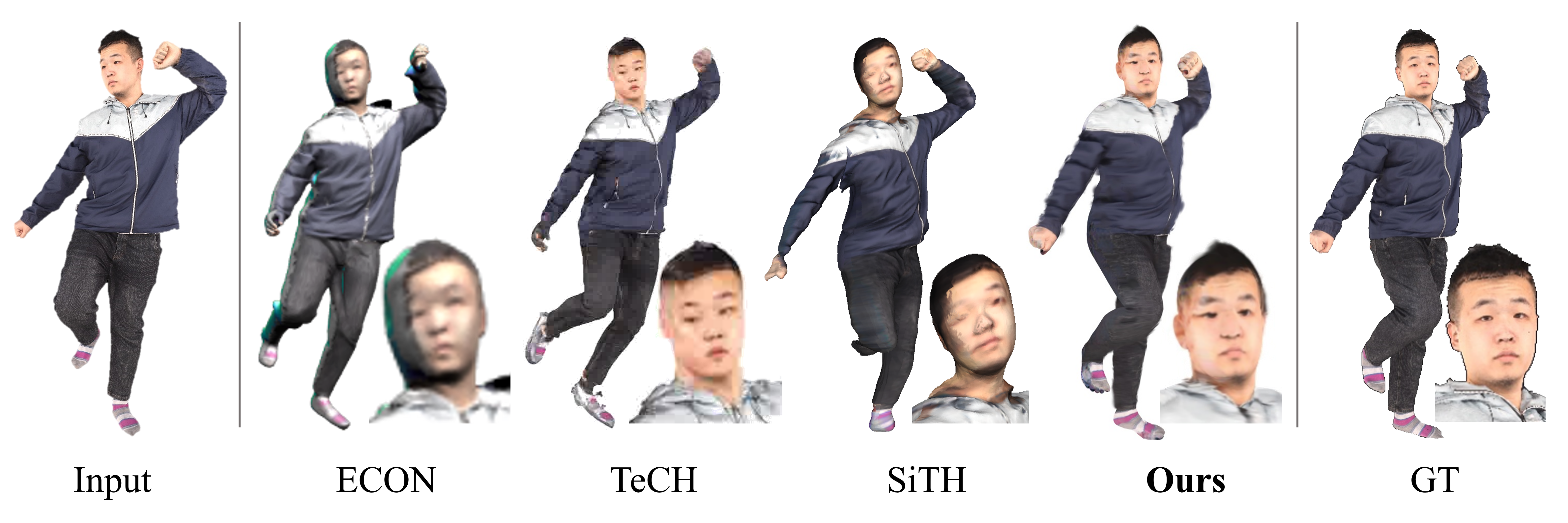}
     \caption{Comparison with single-view reconstruction methods: ECON~\cite{econ}, TeCH~\cite{tech}, and SiTH~\cite{sith}. Our method outperforms the baselines in terms of faithfulness to the given observation. }
   \label{fig:single_view}
\end{figure}

\figref{fig:single_view} shows comparisons with SOTA single-view reconstruction methods that are based on 3D human prior: ECON~\cite{econ}, TECH~\cite{tech}, and SiTH~\cite{sith}. We used their officially released implementation for the comparison.
Our sparse-view work outperforms in terms of accuracy and faithfulness to the observed data, as can be seen in \figref{fig:single_view}.
Also, the single-view methods either require per-subject optimization (ECON, TeCH) or run at relatively slow speed (e.g., ECON 3 min / TeCH 4 hr / SiTH 2 min). On the other hand, ours is a feed-forward method that runs at $4 fps$, which is $\times$480 faster than SiTH.

\subsection{Ablations}
\begin{table}[t]
\centering
\small
\tabcolsep 6pt
\caption{
\small{\textbf{Ablation study on the number of outer scaffolds used.} We trained and tested variants with different numbers of scaffolds that are outside the original SMPL-X surface. The variant with only the base template is denoted as ``0 scaffold''. The performance increase is saturated as more than 5 outer scaffolds are used.} 
}
\vspace{-2mm}

\begin{tabular}{cccc}
       \# Out scaffolds. & PSNR$\uparrow$    & LPIPS$\downarrow$ & FID$\downarrow$    \\
        \toprule
        0 & 22.30     & 145.74    & 84.38 \\
        1 & \bf{22.77}     & 139.16    & 75.66 \\
        2 & 22.28     & 137.65    & 73.54 \\
        3 & 21.87     & 136.38    & 65.19 \\
        4 \bf{(Ours full)} & 21.90     & \bf{133.41}    & \bf{61.67} \\
        5 & 22.13     & 134.73    & 63.80 \\
        6 & 22.09     & 135.52    & 64.81 \\
        \bottomrule
\end{tabular}
\label{tab:supp_num_scaffold}
\end{table}

\begin{figure}[t]
  \centering
  \includegraphics[width=0.7\linewidth]{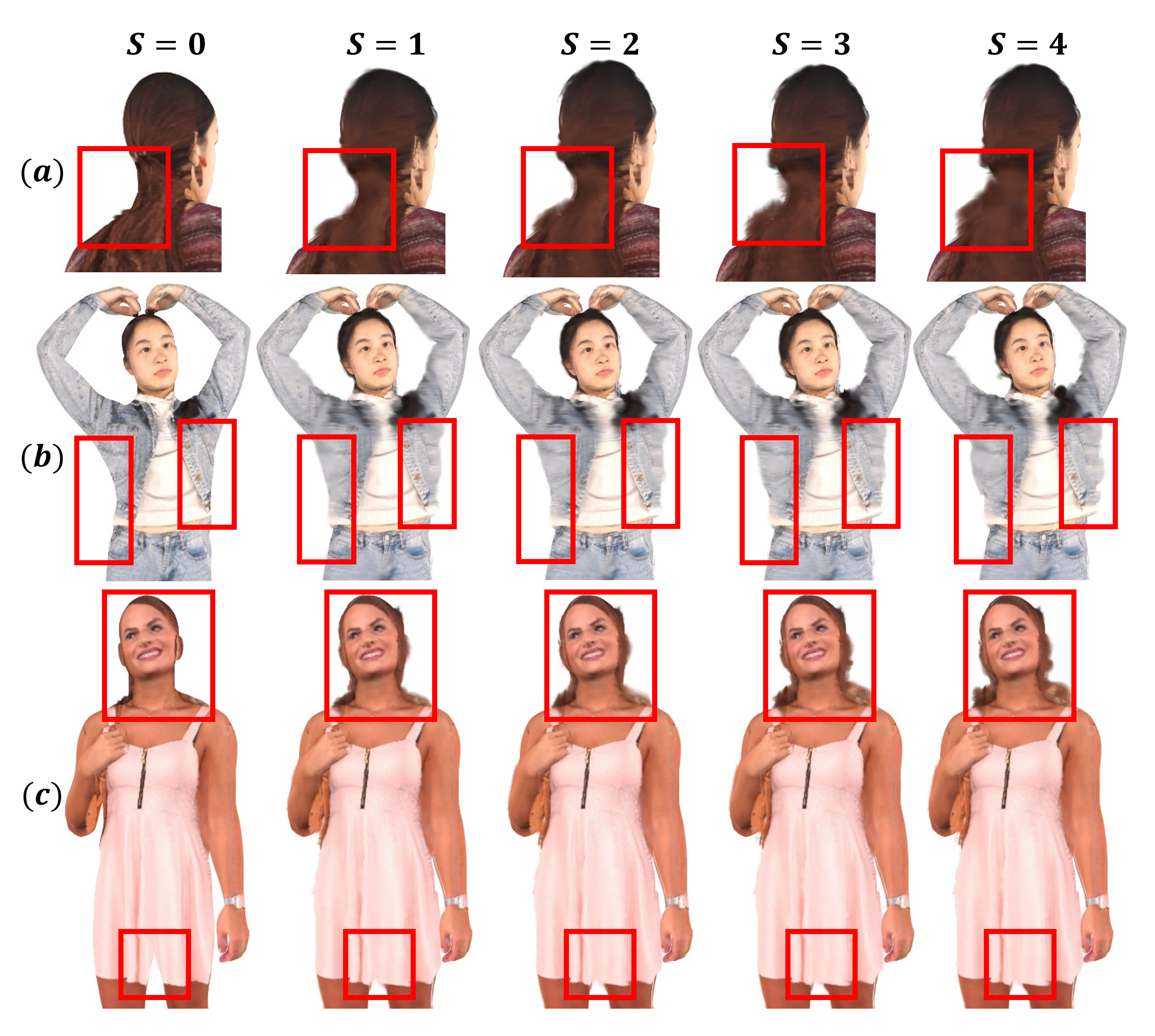}
   \caption{Multi-scaffold helps reconstruct hair and loose clothing. $S$ denotes the number of outer scaffolds. }
   \label{fig:multi_scaffold}
\end{figure}
\begin{table}[t]
\centering
\small
\tabcolsep 7.5pt
\caption{
\small{
\textbf{Ablation study on the supervision.} \xmark/\cmark indicates completely remove/keep the loss supervision. Our $L_1$-only supervision result (a) still outperforms the human template-conditioned NeRF methods NHP and NIA, which are also trained with $L_1$-only supervision. This validates the effectiveness of our proposed multi-scaffold.
}}\label{tab:supp_supervision}
\vspace{-2mm}
\begin{tabular}{lccccccc}
\multicolumn{1}{l}{} & $L_1$ & SSIM & Mask & Multi-view & \multicolumn{1}{c}{PSNR$\uparrow$} & \multicolumn{1}{c}{LPIPS$\downarrow$} & \multicolumn{1}{c}{FID$\downarrow$} \\
\toprule
NHP & \cmark & \xmark & \xmark & \xmark  & \bf{23.32}     & 184.69    & 136.56  \\
NIA & \cmark & \xmark & \xmark & \xmark & 23.20     & 181.82    & 127.30  \\
\midrule
a & \cmark &\xmark & \xmark & \xmark  & 23.05   & 142.57    & 71.97  \\
b & \cmark &\cmark & \xmark & \xmark  & 22.69   & 136.44    & 69.50  \\
c & \cmark &\cmark & \xmark & \cmark & 22.03     & 134.82    & 62.04  \\
\midrule
\bf{Ours full}  & \cmark  & \cmark  & \cmark  & \cmark
    & {21.90}     & \bf{133.41}    & \bf{61.67} \\

\bottomrule
\end{tabular}
\end{table}

\begin{table}[ht]
\centering
\small
\tabcolsep 6pt
\caption{
\small{\textbf{Ablation study on the number of input views at inference.} We trained our model using 3 input views, and tested with different numbers of input views at inference time. The performance improves as more observations are available.} 
}
\vspace{-2mm}

\begin{tabular}{cccc}
       \# Inputs. & PSNR$\uparrow$    & LPIPS$\downarrow$ & FID$\downarrow$    \\
        \toprule
        1 & 20.08     & 152.54    & 99.13 \\
        2 & 21.79     & 132.61    & 78.56 \\
        3 & 21.90     & 133.41    & 61.67 \\
        4 & 22.01     & 133.68    & 53.40 \\
        5 & \bf{22.07}     & \bf{131.80}    & \bf{35.00} \\
        \bottomrule
\end{tabular}
\label{tab:supp_num_input}
\end{table}

\mypara{Ablation on the number of scaffolds.} 
In \tabref{tab:supp_num_scaffold}, we study the impact of number of outer scaffolds. Variants with different number of outer scaffolds are trained and tested. The performance increase is saturated as more than 5 outer scaffolds are used. Therefore, we use 4 outer scaffolds as our final model. In \figref{fig:multi_scaffold}, we show how the number of scaffolds affects the reconstruction of offset details such as hair (a,c) and loose clothing (b,c).

\mypara{Ablation on the supervision.}
\tabref{tab:supp_supervision} shows the impact of different loss supervision employed during training. Note that our variant with $L_1$-only supervision (\tabref{tab:supp_supervision}-a) already outperforms the human template-conditioned generalizable NeRF methods NHP and NIA, which are also trained with $L_1$-only supervision, in terms of perceptual metrics LPIPS and FID. This validates that our gain is not only from the different supervision but also from our proposed multi-scaffold. Our full model that leverages multi-view supervision with $L_1$, SSIM, and mask loss achieves the highest performance on the perception-based metrics. Note that multi-view supervision is possible by leveraging the fast 3D Gaussian splatting.

\mypara{Ablation on the number of input views at inference.}
We trained our model using 3 input views and tested with different number of input views at inference time in \tabref{tab:supp_num_input}. The performance improves as more observations are available. However, note that our performance when only given two views is still comparable to the 3-view results. This demonstrates the effectiveness of our method under sparse view setting.

\mypara{Performance on the randomly selected input views.}
During evaluation, we followed the convention of previous sparse view 3D human reconstruction works~\cite{kwon2021neural, kwon2023neural} that use 3 uniformly distributed inputs.
However, we additionally ran the evaluations given 3 random views 10 times and computed the mean metrics. We verified that the performance difference between the uniformly and randomly sampled inputs is minimal -- PSNR is $1.5\%$, and LPIPS is 0.3\%.

\subsection{Runtime at inference}
Our GHG runs at $4fps$ for rendering a single $1K$ ($1024 \times 1024$) image on a single NVIDIA RTX A4500 GPU. However, note that inpainting network takes most of our runtime ($74\%$). Without the inpainting network, ours runs at $15fps$. More efficient inpainting model can be explored to further reduce the runtime.

The detailed breakdown of runtime is as follows.
Our pipeline can be divided into three stages: (1) constructing multi-scaffold (2) Gaussian parameter map generation (3) rasterization.
\textbf{(1) Constructing multi-scaffold:} RGB map for each scaffold is aggregated on the UV space of human template. Our inpainting network inpaints the missing regions of the innermost scaffold RGB map in 180.89 ms. 
\textbf{(2) Gaussian parameter map generation:} Multi-Gaussian parameter maps are generated in 57.97 ms.
\textbf{(3) Rasterization:} Rasterization takes 5.78 ms. In total, GHG takes 244.65 ms to render a single $1K$ image.

We would like to highlight that our method runs faster than the sparse-view generalizabl human NeRF methods NHP and NIA ($0.01fps$ to render a single 1K image) while outperforming their visual quality.

\section{Implementation details} 
\begin{figure}[h]
    \centering
    \includegraphics[width=\textwidth]{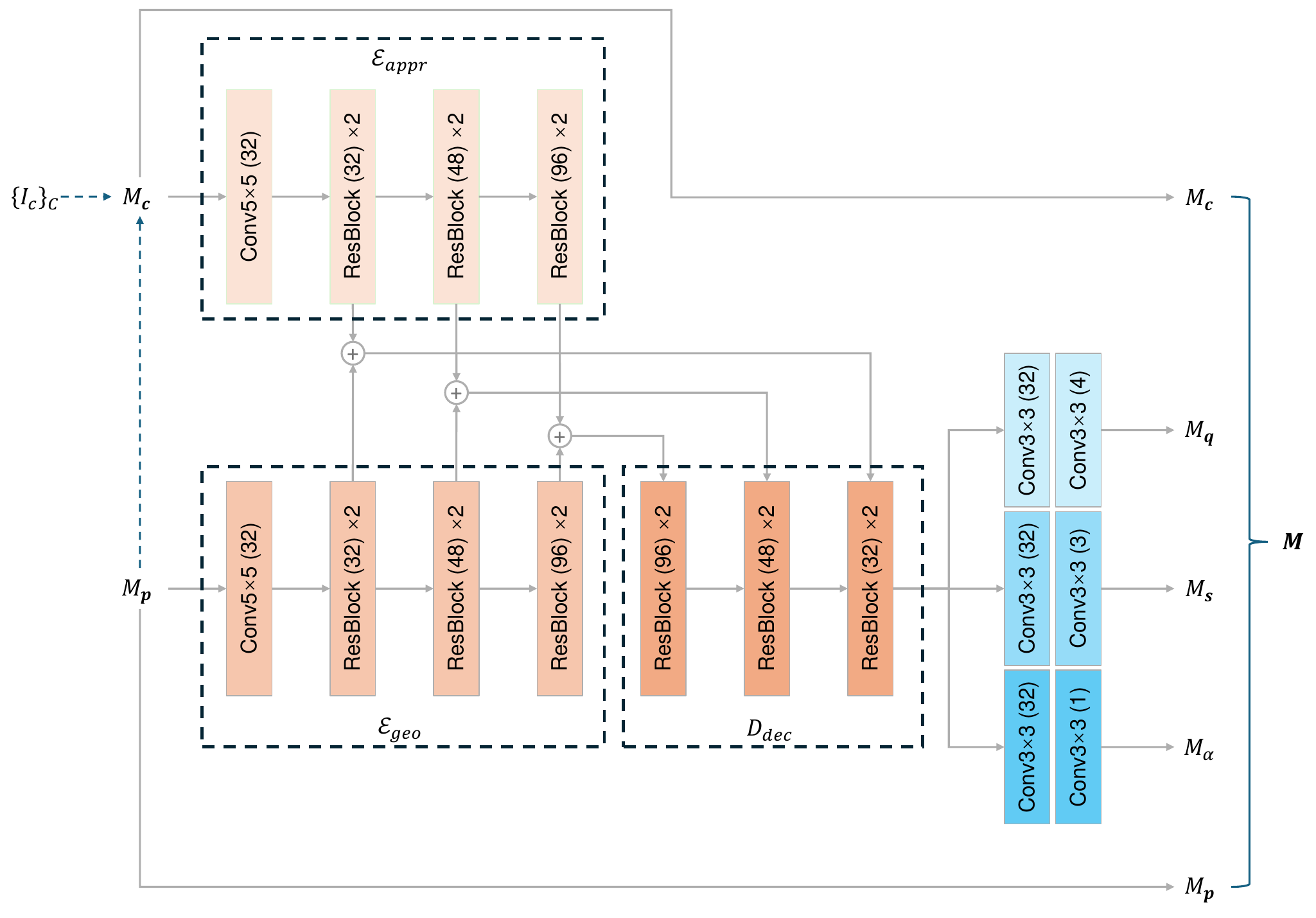}
    \caption{\textbf{Network architecture for Gaussian parameter map generation.}}
    \label{fig:supp_gaussian_network}
\end{figure}

\label{sec:reproducibility}
\subsection{Gaussian parameter map generation}
The architecture design of our Gaussian parameter map generation network is presented in \figref{fig:supp_gaussian_network}. 
Our network is composed of two encoders $\mathcal{E}_\text{appr}$, $\mathcal{E}_\text{geo}$ and one decoder $\mathcal{D}_\text{dec}$. The feature maps extracted by $\mathcal{E}_\text{appr}$ and $\mathcal{E}_\text{geo}$ are added together before being fed into $\mathcal{D}_\text{dec}$. Moreover, $M_\mathbf{s}$ and $M_{\alpha}$ are sent into \textit{Softplus} and \textit{Sigmoid} activation layers, respectively, after the convolution layers. Note that in the figure, the number following each layer name and sitting in the bracket denotes its output channel size.

\begin{figure}[h]
\centering
\def\arraystretch{0.5}
\includegraphics[width=0.9\linewidth]{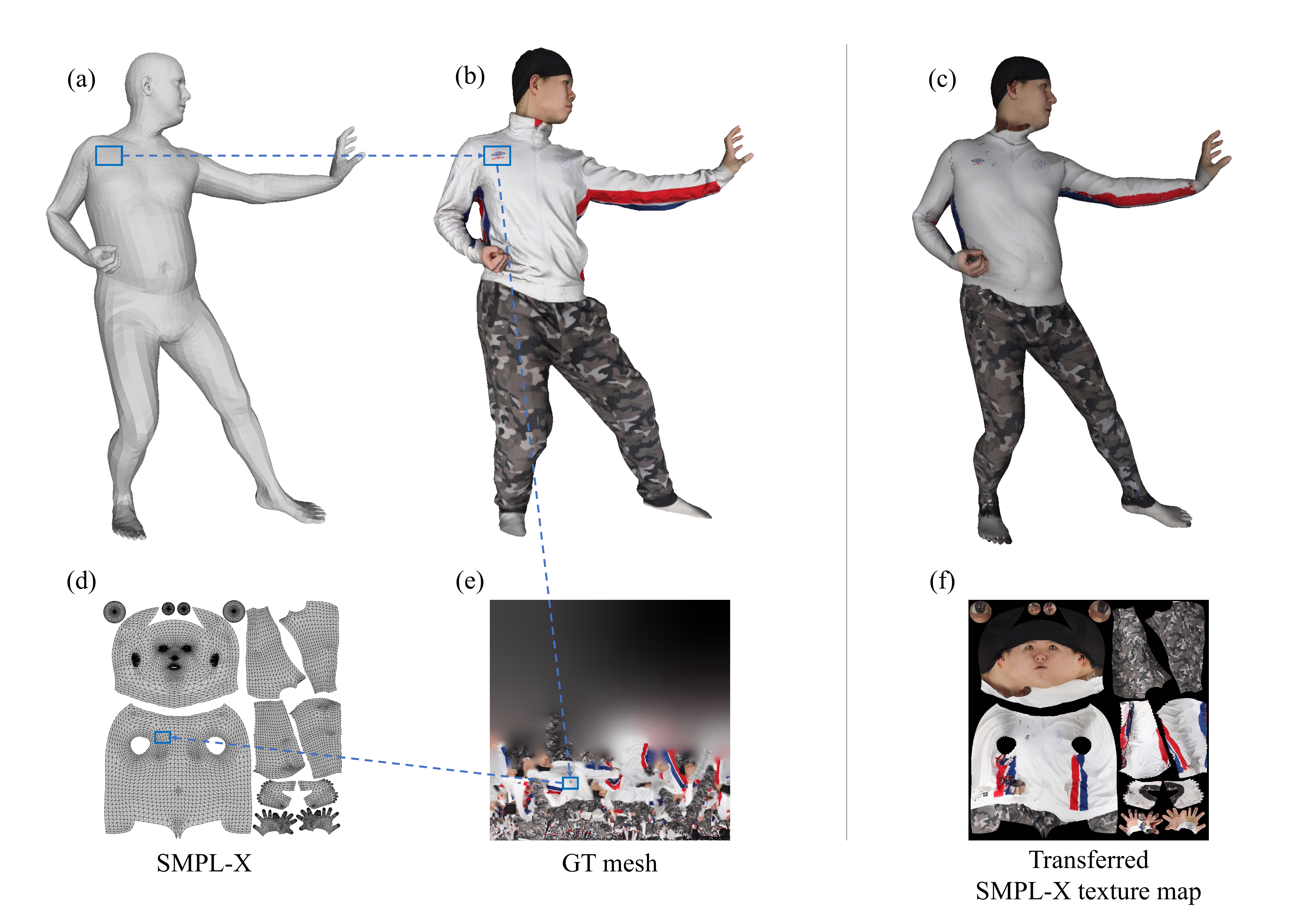}
 
\caption{\small \textbf{Illustration of texture transfer on to the SMPL-X UV space.} For each point on the SMPL-X model (a), the nearest point on the scanned mesh (b) is found. Then, we get the corresponding position of this point on the scan's UV map (e), which will be mapped to the matching location on the SMPL-X's UV map (d). Resulting on the transferred texture map (f) and the colored mesh (c). 
}
\label{fig:supp_inpaint_gt}
\vspace{-2mm}
\end{figure}

\subsection{Inpainting}

\mypara{Pseudo ground truth generation}
To create the pseudo ground truth texture map on the SMPL-X UV space, we follow the approach proposed in Lazova et al~\cite{lazova2019360}. The process is illustrated in \figref{fig:supp_inpaint_gt}.
For each point on the SMPL-X model, we identify the nearest point on the scanned object. Next, we determine the corresponding position of this point on the scan's UV map. We then transfer the color from this position on the scan's UV map to the corresponding location on the SMPL-X's UV map.

\begin{figure}[h]
    \centering
    \includegraphics[width=0.99\textwidth]{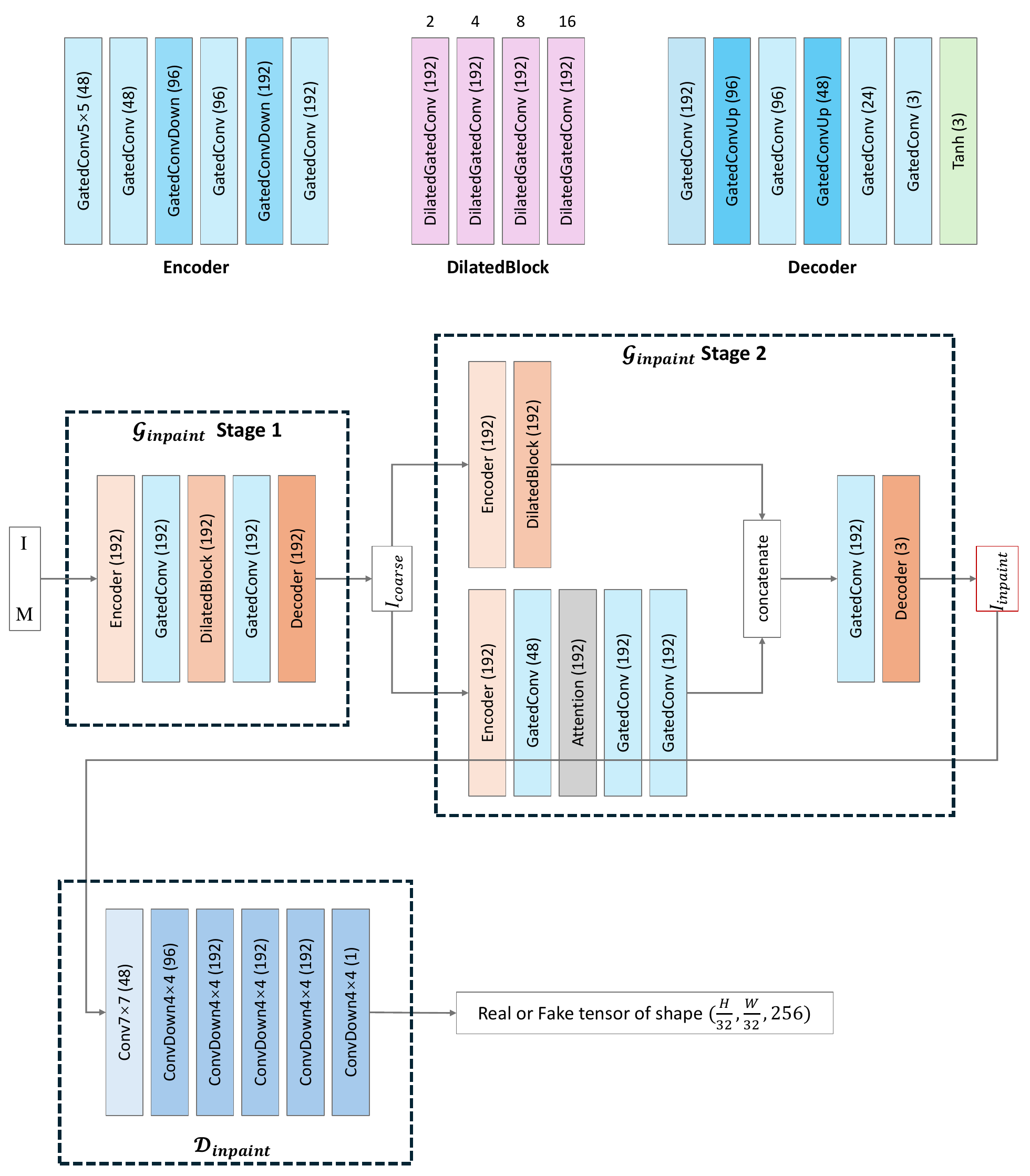}
    \caption{\textbf{Inpainting network.}}
    \label{fig:supp_inpaint_network}
\end{figure}

\mypara{Network architecture}
\figref{fig:supp_inpaint_network} shows the inpainting module architecture. The inpainting network follows the DeepFillv2 design~\cite{yu2018free}. 
The inpainting network is composed of a generator $\mathcal{G}_\text{inpaint}$ and a discriminator $\mathcal{D}_\text{inpaint}$. In the generator, all convolutions are gated convolutions with a kernel size of $3\times 3$ if not specified, where \textit{GatedConv}, \textit{DilateGatedConv}, \textit{GatedConvDown}, \textit{GatedConvUp} have a stride of 1, 1, 2, 0.5, respectively. The four \textit{DilateGatedConv} layers in \textit{DilatedBlock} have a dilation of 2, 4, 8, 16, respectively. The \textit{Attention} layer is a self-attention layer. 
In the discriminator, all convolutions are common 2D convolutions, where \textit{Conv}, \textit{ConvDown} have a stride of 1, 2, respectively. Besides, all convolution layers are followed by ELU activation. Note that in the figure, the number following each layer name and sitting in the bracket denotes its output channel size. 

\end{document}